\documentclass[10pt,conference]{IEEEtran}
\IEEEoverridecommandlockouts
\usepackage{cite}
\usepackage{amsmath,amssymb,amsfonts}
\usepackage{textcomp}
\usepackage{xcolor}
\usepackage{amsmath}
\usepackage[linesnumbered,ruled,vlined]{algorithm2e}
\usepackage{hyperref}
\hypersetup{hidelinks}
\usepackage{multirow}
\usepackage{bm}
\usepackage[normalem]{ulem}

\usepackage{tikz}
\usetikzlibrary{shapes}
\usetikzlibrary{positioning}
\usetikzlibrary{plotmarks}
\usetikzlibrary{patterns}
\usetikzlibrary{fit,calc}
\usepackage{pgfplots}
\usetikzlibrary{arrows.meta}
\usepackage[customcolors,shade]{hf-tikz}
\usepackage{microtype}
\usepackage{graphicx}
\usepackage{subcaption} 
\usepackage{enumitem}
\usepackage{booktabs}
\usepackage{adjustbox}
\usepackage{xspace}
\usepackage{multirow}
\usepackage{thmtools, thm-restate}
\usepackage{siunitx}
\usepackage[tablename=Table]{caption}
\usepackage{algorithmic}
\usepackage{balance}
\usepackage{url}

\ifCLASSOPTIONcompsoc

\else

\fi

\def\BibTeX{{\rm B\kern-.05em{\sc i\kern-.025em b}\kern-.08em
    T\kern-.1667em\lower.7ex\hbox{E}\kern-.125emX}}

\begin{document}

\title{Edge-Cloud Collaborative Motion Planning for Autonomous Driving with Large Language Models}
\author{\IEEEauthorblockN{
Jiao Chen\IEEEauthorrefmark{1}, 
Suyan Dai\IEEEauthorrefmark{1}
Fangfang Chen\IEEEauthorrefmark{1}, 
Zuohong Lv\IEEEauthorrefmark{1}
and Jianhua Tang\IEEEauthorrefmark{1}\IEEEauthorrefmark{2}}
    \IEEEauthorblockA{\IEEEauthorrefmark{1} Shien-Ming Wu School of Intelligent Engineering, South China University of Technology, China}
    \IEEEauthorblockA{\IEEEauthorrefmark{2} Pazhou Lab, Guangzhou, China}
    \IEEEauthorblockA{\{202110190459, 202066200091, wifannychen59, 202220159664\}@mail.scut.edu.cn, jtang4@e.ntu.edu.sg}
\thanks{The corresponding author is Jianhua Tang.}
\thanks{Pre-print, manuscript submitted to IEEE.}
}

\maketitle
\begin{abstract}
Integrating large language models (LLMs) into autonomous driving enhances personalization and adaptability in open-world scenarios. However, traditional edge computing models still face significant challenges in processing complex driving data, particularly regarding real-time performance and system efficiency. To address these challenges, this study introduces EC-Drive, a novel edge-cloud collaborative autonomous driving system with data drift detection capabilities. EC-Drive utilizes drift detection algorithms to selectively upload critical data, including new obstacles and traffic pattern changes, to the cloud for processing by GPT-4, while routine data is efficiently managed by smaller LLMs on edge devices. This approach not only reduces inference latency but also improves system efficiency by optimizing communication resource use. Experimental validation confirms the system’s robust processing capabilities and practical applicability in real-world driving conditions, demonstrating the effectiveness of this edge-cloud collaboration framework. Our data and system demonstration will be released at \url{https://sites.google.com/view/ec-drive}.
\end{abstract}

\section{Introduction}
As intelligent transportation and autonomous driving technologies rapidly advance, the motion planning system, as a critical component, faces increasingly complex environments and diverse challenges. Traditional motion planning methods often rely on fixed algorithms and models, making it difficult to fully address the dynamic changes in traffic conditions and the personalized needs of drivers \cite{xiong2020communication}.

Integrating large language models (LLMs) into autonomous vehicles not only enables artificial intelligence systems to control the driving process but also significantly enhances the system's personalization and adaptability. By understanding natural language commands, LLMs can dynamically adjust driving strategies to meet the personalized preferences of drivers or passengers, thereby improving the overall driving experience. Moreover, the integration of LLMs allows autonomous systems to better handle complex and dynamic open-world scenarios, making them more flexible in addressing diverse driving tasks. 

The Transformer, originally designed for sequential data, has achieved state-of-the-art performance in natural language processing, driving the development of LLMs  \cite{achiam2023gpt, touvron2023llama}. 
These models pretrain Transformer architectures (encoder, encoder-decoder, and decoder) on vast corpora to capture extensive language statistics. Pretrained LLMs can be fine-tuned for specialized downstream tasks. 
The Vision Transformer (ViT) \cite{dosovitskiy2020image} applies the Transformer to image tasks, converting images into sequences of patches that the Transformer can process. CLIP \cite{radford2021learning}, a multimodal model that matches textual descriptions with images, demonstrates strong transfer capabilities in many image classification tasks. Utilizing pretrained LLMs as a framework for multimodal tasks leverages their text generation capabilities, which is crucial for the question-answering tasks in our research. However, despite their impressive performance in many tasks, deploying these large models with typically over a billion parameters for real-time applications remains challenging. \looseness=-1

\begin{figure}[t]
    \centering
    \includegraphics[width=0.9\linewidth]{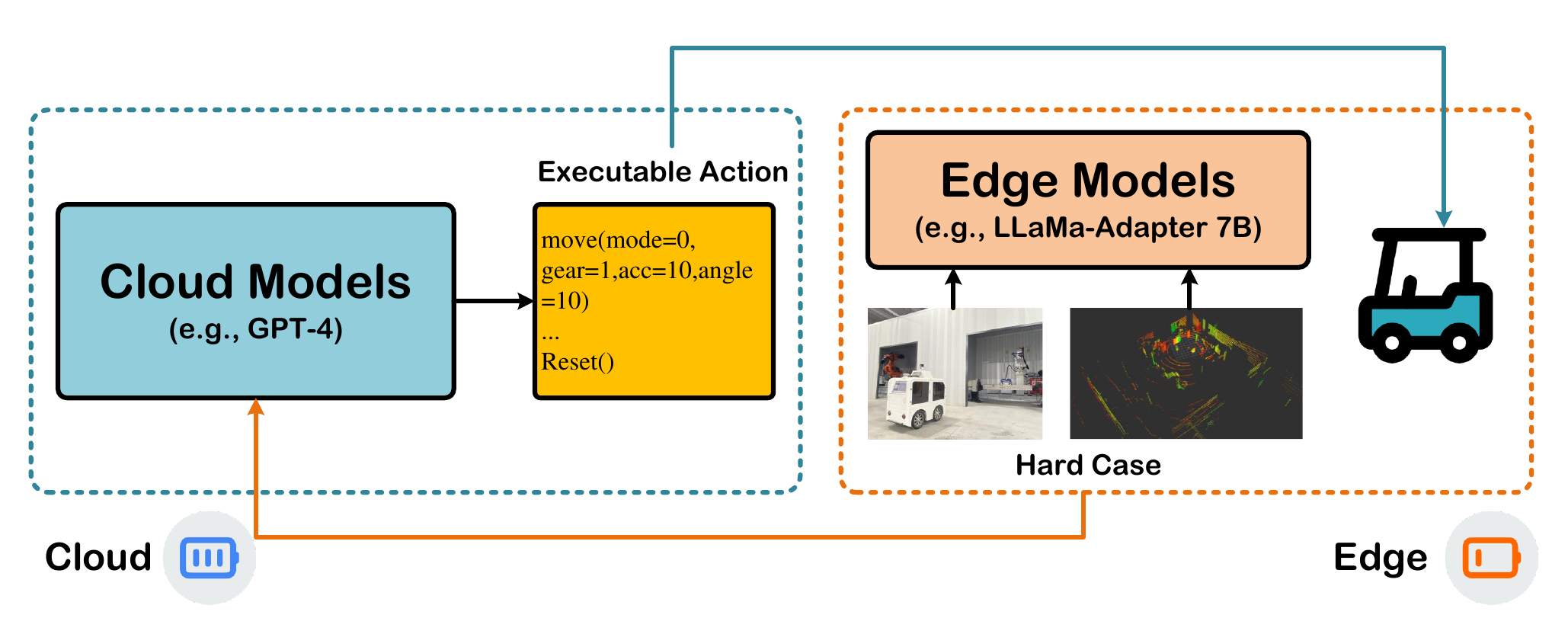}
    \caption{Architecture of the EC-Drive system. LLM-based motion planning is performed on edge devices within the vehicle, while complex inference tasks are offloaded to the cloud, which has larger models and more extensive resources.
    }
    \label{fig:ec}
    \vspace{-0.15in}
\end{figure}

Although autonomous driving systems primarily rely on visual features, incorporating linguistic features can enhance system interpretability and aid in identifying new traffic situations. This advantage has sparked interest in integrating multimodal data to train language models as autonomous driving agents.
DriveGPT4 \cite{xu2023drivegpt4} employs LLaMA as the backbone LLM, with CLIP as the visual encoder, using traffic scene videos and prompt texts as inputs to generate responses and low-level vehicle control signals.
DriveMLM \cite{wang2023drivemlm} utilizes multi-view images, LiDAR point clouds, traffic rules, and user instructions from a real simulator to perform closed-loop driving. This multimodal model is constructed with LLaMA and ViT as the image processor.
GPT-Driver \cite{mao2023gpt} reframes motion planning as a language modeling task, using GPT-3.5 to represent the planner’s inputs and outputs as language tokens. 

However, these models utilize LLMs with over a billion parameters (such as GPT-3.5 \cite{achiam2023gpt} and LLaMA \cite{touvron2023llama}) and expensive image encoders (such as CLIP and ViT), making them suitable mainly for latency-insensitive offline scenarios rather than latency-critical online scenarios. Recently, collaboration between large and small language models has garnered significant attention \cite{zhang2024fast}. Inspired by dual-process cognitive theory, various methods can be integrated into a unified framework.

Our primary insight is to use data drift detection algorithms to upload a small number of difficult samples (e.g., new obstacles, changes in traffic patterns) to the cloud for processing by larger-scale models (e.g., GPT-4), while most samples are handled by smaller parameter LLMs at the edge. This approach, illustrated in Fig.~\ref{fig:ec}, ensures low inference latency while improving the handling of dynamic environments. This method has potential applications in remote assistance for autonomous vehicles, enabling them to navigate complex and evolving scenarios more effectively. Our main contributions are as follows:

$\bullet$ We propose a novel edge-cloud collaborative autonomous driving system, EC-Drive, equipped with data drift detection capabilities. This efficient framework utilizes data drift detection algorithms to selectively upload a small number of challenging samples (e.g., new obstacles, changes in traffic patterns) to the cloud for processing by GPT-4, while most of the data is managed by smaller parameter LLMs on edge devices. This approach ensures low inference latency while effectively addressing the challenges of complex environments.

$\bullet$ We introduce a multimodal approach that integrates linguistic features with traditional visual data, enhancing the interpretability and decision-making capabilities of autonomous driving systems. This integration allows the system to better understand and respond to new traffic situations, improving adaptability and safety.

$\bullet$ Detailed experimental validation demonstrates the system’s robust processing capabilities and its potential applicability in real-world driving scenarios, highlighting the practical advantages and feasibility of the proposed edge-cloud collaborative framework. \looseness=-1

\section{Related Works}
This section reviews motion planning methods and the practical application of LLMs in autonomous driving, focusing on their strengths and challenges in complex traffic environments.

\subsection{Motion Planning in Autonomous Driving}
Autonomous driving utilizes various motion planning strategies for efficient vehicle navigation.
(1) Rule-based method: This approach generates paths based on predefined rules that account for environmental constraints like road geometry and traffic signals \cite{thrun2006stanley}. While simple and efficient, it is rigid and struggles to adapt to unexpected changes.
(2) Optimization-based method: Optimization algorithms compute optimal trajectories by minimizing a cost function considering factors such as time, energy, safety, and comfort \cite{xiong2020communication}. Though precise, these methods are computationally intensive and may not suit real-time decision-making.
(3) Learning-based method: This approach uses machine learning to adapt to dynamic environments by learning from past data \cite{hu2023planning}. Deep neural networks and reinforcement learning provide adaptability but require significant data and resources, often struggling with rare or novel scenarios.

\subsection{Large Models}
Large models (LMs) based on the Transformer, such as Large Language Models \cite{touvron2023llama,achiam2023gpt}, vision models \cite{dosovitskiy2020image,radford2021learning}, time series models \cite{zhou2024one, he2024continual}, and multimodal models \cite{liu2024visual}, have gained widespread attention due to their unique advantages. With billions to trillions of parameters, these models accumulate extensive knowledge through pre-trained on large datasets, significantly advancing the automation and diversification of data processing while reducing reliance on human expertise. Such capabilities have attracted broad interest in the industrial sector, fostering numerous studies targeting industrial intelligence.

The collaboration between large and small language models garners considerable attention. Inspired by dual-process cognitive theory, various methods can be integrated into a unified framework. Research indicates that the essential difference between large and small models lies in the control of uncertainty in next token predictions during the decoding process, and it highlights that collaborative interactions between models are most critical at the beginning of the generation process \cite{zhang2024fast}.

\subsection{Motion Planning with LLMs.}
In recent years, significant progress has been made in the application of LLMs in the field of autonomous driving.
Utilizing LLMs to enhance decision-making processes in autonomous vehicles has the potential to transform their operational methods. This approach offers personalized assistance, facilitates continuous learning, and improves decision intelligence \cite{cui2024drive}.
PlanAgent\cite{zheng2024planagent} is a multimodal large language model-based autonomous motion planning agent system that enhances environmental understanding through Bird's Eye View (BEV) and lane-graph-based textual descriptions. It introduces a hierarchical Chain of Thought (CoT) \cite{wei2022chain} to guide the MLLM in generating planner code.
Hu \textit{et al.} \cite{hu2024agentscodriver} propose an LLM-driven collaborative driving framework for multiple vehicles, featuring lifelong learning capabilities. It allows different driving agents to communicate with each other, facilitating collaborative driving in complex traffic scenarios.
DiLu \cite{wen2023dilu} is the first framework to leverage knowledge-driven capabilities in autonomous driving decision-making. It combines reasoning and reflection modules, enhancing the capabilities of LLMs, enabling them to apply knowledge and perform causal reasoning in the autonomous driving domain.
TrafficGPT \cite{zhang2024trafficgpt} reveals the application potential of large language models in the smart transportation domain. These models possess the capability to view and process traffic data, providing profound decision support for urban traffic system management. Additionally, they assist in human decision-making during traffic control, demonstrating their practicality and efficacy in traffic management.

LimSim++ \cite{fu2024limsim++} is an open-source evaluation platform specifically designed for the research of autonomous driving  with LVLMs, supporting scenario understanding,  decision-making, and evaluation.
DriveLM \cite{sima2023drivelm} introduces datasets using nuScenes and CARLA, presenting a vision-language models based baseline approach that concurrently addresses Graph visual question answering and end-to-end driving. The experiments showcased Graph visual question answering as a simple and principled framework for scene reasoning.
CODA-LM \cite{li2024automated} demonstrates that even the most advanced autonomous driving perception systems struggle with handling complex road corner cases. \looseness=-1

\begin{figure}[!t]
    \centering
    \includegraphics[width=0.9\linewidth]{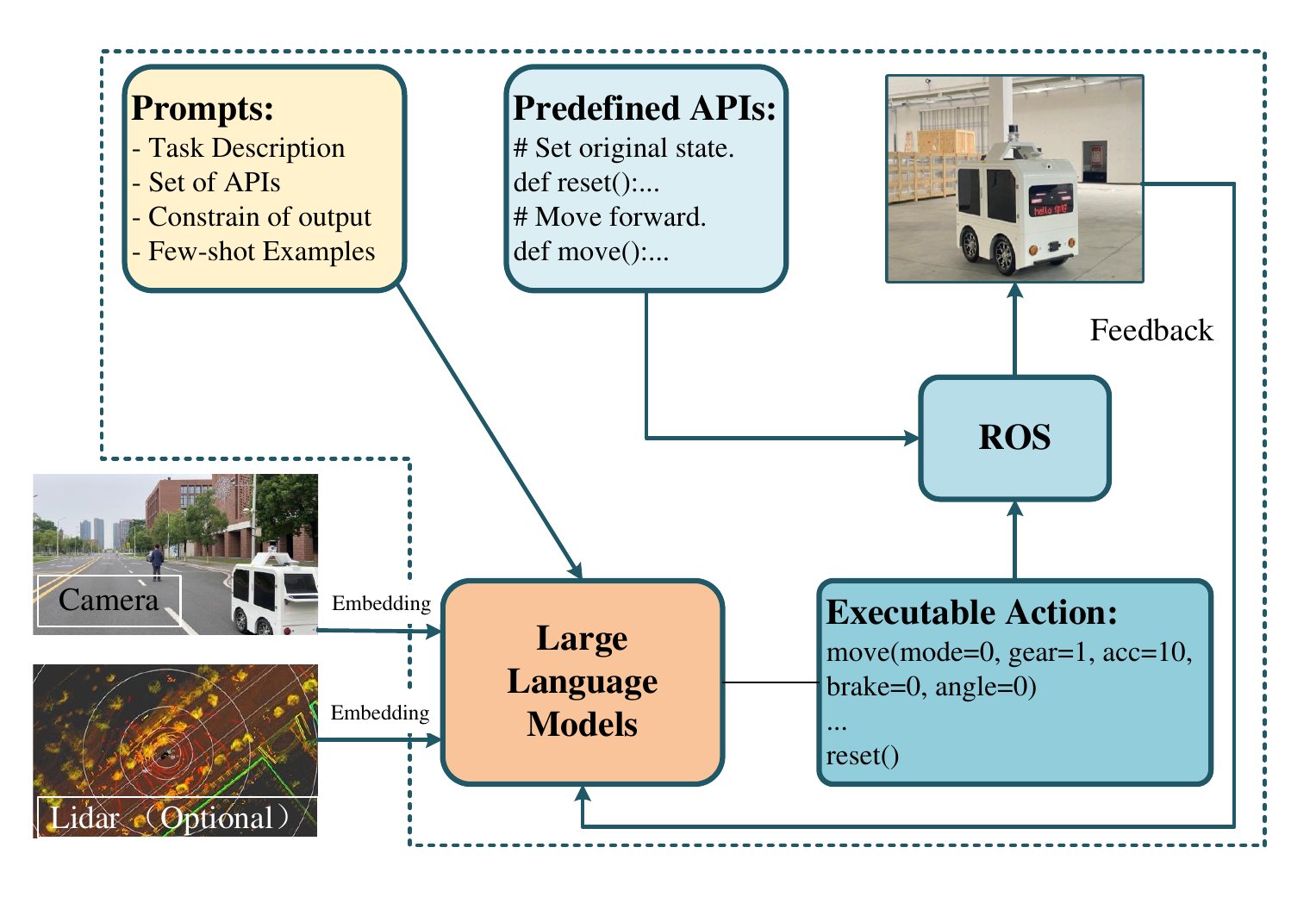}
    \caption{Motion planning process on the edge through large language models, utilizing vision and LiDAR data for real-time decision-making and execution. ROS stands for Robot Operating System, which is used to execute actions and provide feedback on the execution results.}
    \label{fig:local}
    \vspace{-0.15in}
\end{figure}

\section{Edge-Cloud Collaborative Motion Planning for Autonomous Driving}
In this section, we elaborate on the methodologies and technologies employed in the EC-Drive system, emphasizing the use of edge and cloud models as well as the collaborative process between them. This approach ensures efficient and safe decision-making, even in complex driving environments.

\subsection{Problem Statement}
In edge-cloud collaborative intelligent driving systems, we deploy small-scale LLMs on edge devices for real-time motion planning and large-scale LLMs on the cloud to provide efficient support. Edge devices, when processing real-time driving data, may encounter distribution shifts or decreased model confidence due to natural variations (such as changes in lighting or weather) or sensor degradation, which can affect model performance.

Two primary scenarios necessitate the request for support from large models in the cloud: 

(1) When the vehicle encounters new or previously unseen objects or situations, increasing decision-making complexity, and the edge model may be insufficient for accurate inference.

(2) For instance, visual obstructions or lighting variations may reduce the accuracy and reliability of edge model predictions. Under such circumstances, leveraging large models in the cloud for deeper analysis can enhance system performance and safety. \looseness=-1

\begin{figure}[!t]
    \centering
    \includegraphics[width=1\linewidth]{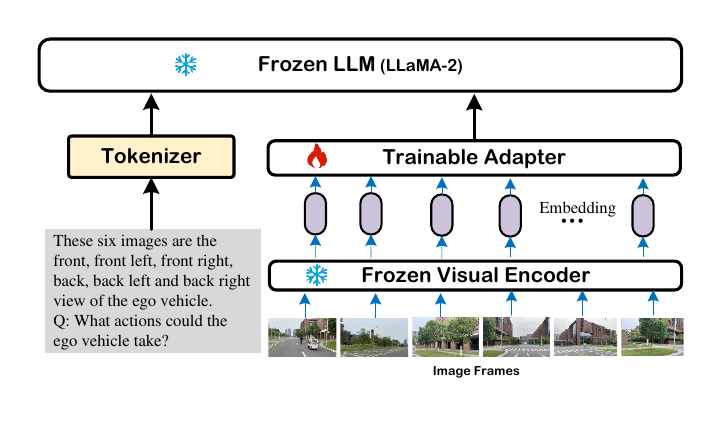}
    \caption{Instruction tuning of pretrained LLaMA2 models for autonomous driving, using multi-view images and prompt for efficient adaptation to specific driving scenarios.}
    \label{fig:instruction tuning}
    \vspace{-0.15in}
\end{figure}

\begin{figure*}
    \centering
    \includegraphics[width=0.9\linewidth]{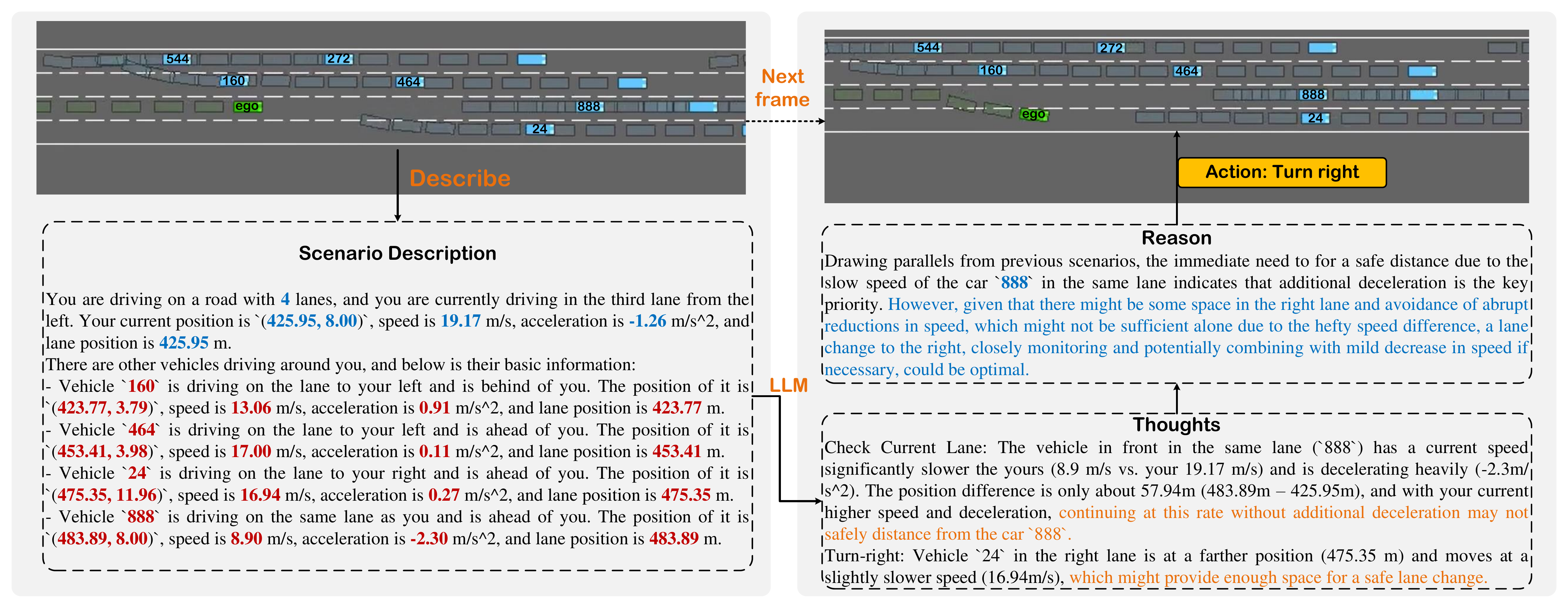}
    \caption{Edge model performs step-by-step reasoning and decision making in a complex traffic environment}
    \label{fig:case study 1}
    \vspace{-0.15in}
\end{figure*}

\subsection{System Architecture}
The proposed system architecture, illustrated in Fig.~\ref{fig:local}, integrates edge and cloud components to enhance the overall performance of autonomous driving systems. The vehicle employs small-scale LLMs, fine-tuned using instruction-based approaches as shown in Fig.~\ref{fig:instruction tuning}, to manage routine driving tasks and process real-time sensor data for immediate decision-making. \looseness=-1

\textbf{Edge Models}: 
We employ LLaMA-Adapter \cite{gao2023llama}, a parameter-efficient tuning mechanism based on the LLaMA language model. LLaMA-Adapter is specifically designed for scenarios where computational resources are constrained, such as autonomous driving. It introduces small, zero-initialized attention modules, which are fine-tuned to adapt to new tasks without modifying the entire pre-trained model. This approach minimizes the additional computational overhead, making it ideal for real-time motion planning on edge devices. The model processes real-time sensor data, including text, vision and LiDAR inputs, to make preliminary driving decisions under normal conditions. Pre-print, manuscript submitted to IEEE.

\textbf{Cloud Models}: In the cloud, large-scale LLMs such as GPT-4 offer advanced computational power for handling more complex and dynamic driving scenarios. Real-time data from various onboard sensors, including cameras, LiDAR, and radar, is collected and preprocessed to extract pertinent features. This preprocessing converts the raw sensor data into a structured format that is amenable to model inference. The processed data is then input into the edge model for initial inference, facilitating efficient and timely driving decisions under varying conditions. \looseness=-1

\textbf{Edge-Cloud Collaboration Workflow:} Inspired by \cite{10419797}, we utilize the Alibi Detect library \cite{alibi-detect} to monitor edge model performance. If anomalies or low-confidence predictions are detected, the system flags those instances and uploads the data to the cloud. The cloud model then performs detailed inference to generate optimized decisions, which are integrated with the edge model’s outputs to update the vehicle’s driving plan, ensuring safe and efficient operation.

Let $\textbf{\text{x}}$ denote the preprocessed driving data, and the inference result of the edge model is $a = f_{\text{edge}}(\textbf{\text{x}})$, where $f_{\text{edge}}$ represents the edge model. We use the Alibi Detect library to perform anomaly detection. If the prediction result $p = cd.\text{predict}(\textbf{\text{x}})$ indicates the presence of data drift or low confidence, cloud model support is requested, resulting in an enhanced decision $a' = f_{\text{cloud}}(\textbf{\text{x}})$, where $f_{\text{cloud}}$ represents the cloud model. The overall process is shown in Algorithm~\ref{alg: edgedrive}.

\begin{algorithm}[ht]
\caption{EC-Drive: Edge-Cloud Collaborative Motion Planning}
\label{alg: edgedrive}
\begin{algorithmic}[1]
\STATE \textbf{Initialization:}
\STATE Deploy small-scale LLM on edge for motion planning
\STATE Deploy large-scale LLM on cloud for motion planning
\STATE Initialize Alibi Detect detector $cd$ with reference data

\STATE \textbf{Main Process:}
\WHILE{driving}
    \STATE Collect driving data $\mathcal{D}$
    \STATE Preprocess data: $\textbf{\text{x}} \gets \text{preprocess}(\mathcal{D})$
    \STATE Perform edge model inference on $\textbf{\text{x}}$ and execute decision $a = f_{\text{edge}}(\textbf{\text{x}})$
    \STATE \textbf{Performance Monitoring:}
    \STATE $p \gets cd.\text{predict}(\textbf{\text{x}})$
    \IF {$p$ indicates drift \OR low confidence of edge model}
        \STATE Request cloud model support and execute enhanced decision $a' = f_{\text{cloud}}(\textbf{\text{x}})$
    \ENDIF
\ENDWHILE
\end{algorithmic}
\end{algorithm}

\section{Experiments}
In this section, we present experimental investigations into the real-time operational capabilities of autonomous driving systems using LLMs under different computational paradigms: Edge, Cloud, and Edge-Cloud Collaborative scenarios. Each subsection details distinct approaches and methodologies—ranging from handling in-vehicle data processing at the edge to leveraging cloud computational power for intensive data analysis and decision-making. This comparative study aims to highlight the efficiency, scalability, and reliability of each model under varied driving conditions and their implications on autonomous driving technologies.

\subsection{Driving on Edge}

\textbf{Scene Description:}
We transcribe the current driving scene into descriptive text, including the current speed, acceleration, position of the ego vehicle, and information about surrounding vehicles. For example, the ego vehicle is traveling in the rightmost lane of a four-lane road at a speed of 25.0 m/s, with an acceleration of 0.0 m/s², and its lane position is 361.18 m. The information for other vehicles includes their speed, acceleration, and relative position, such as vehicle 496 in the left lane, ahead by 372.81 m, traveling at a speed of 21.2 m/s, with an acceleration of 0.2 m/s².

\textbf{Reasoning and Thinking:}
The scene description is embedded into vectors and input into the LLaMA-Adapter. Using CoT techniques, LLaMA-Adapter generates sequential reasoning logic and performs step-by-step logical reasoning. For instance, it first assesses whether the vehicle can accelerate. If not, it evaluates the safety of maintaining the current speed. If necessary, it further evaluates the possibility and safety of changing lanes.

\textbf{Decision Making:}
As shown in Fig.~\ref{fig:local}, the system decodes the final decision from the LLM's response and translates it into corresponding vehicle actions, following the outlined process.

As shown in Fig~\ref{fig:case study 1}, we demonstrate how the LLM performs step-by-step logical reasoning and decision-making in a complex traffic environment.

\subsection{Driving on Cloud}
\begin{figure}[!t]
    \centering
    \includegraphics[width=0.9\linewidth]{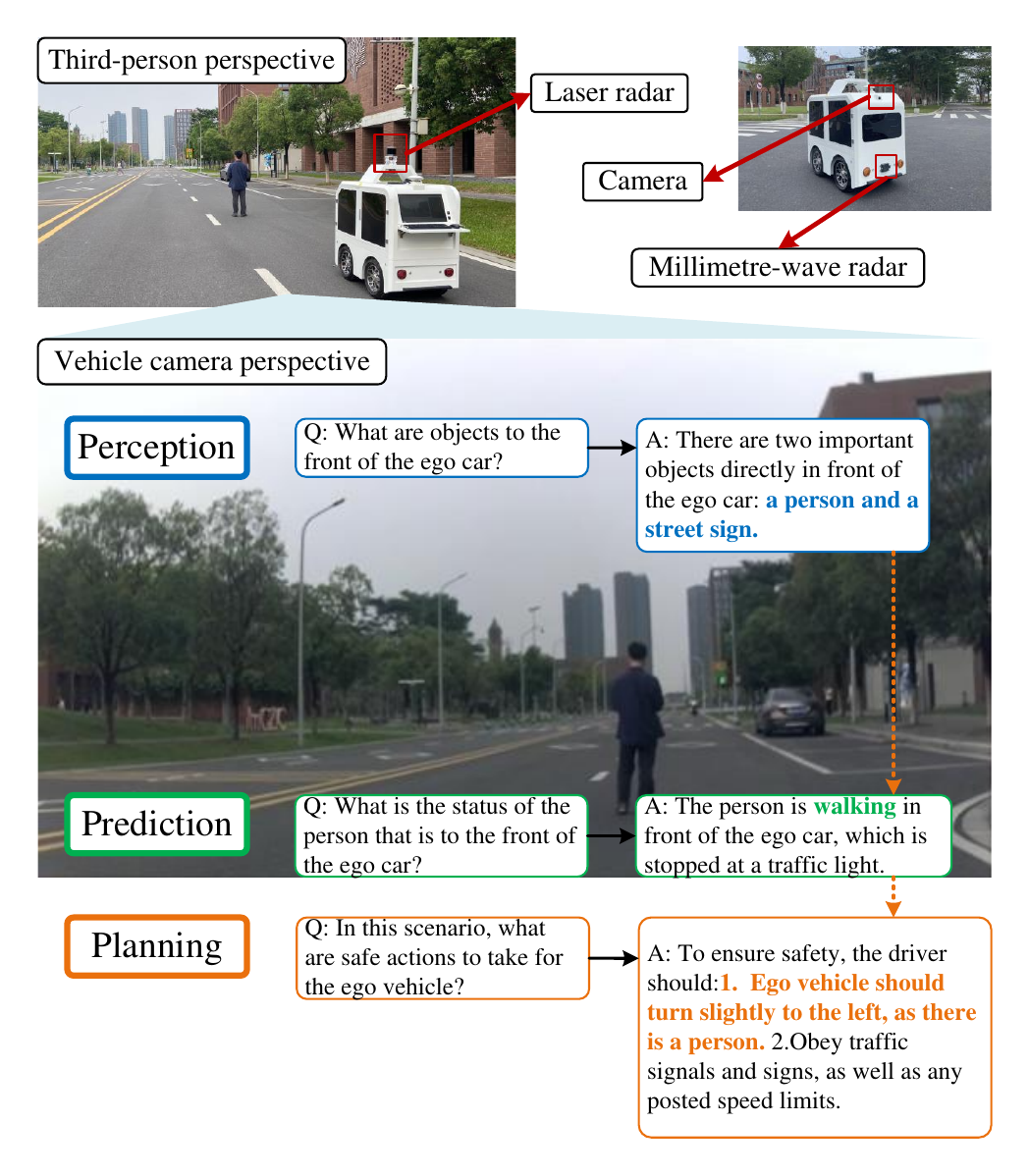}
    \caption{The cloud model addresses incremental driving scenarios, and the yellow dotted line shows the logical dependencies between stages.}
    \label{fig:driving on the cloud}
\vspace{-0.15in}
\end{figure}

As depicted in Fig.~\ref{fig:driving on the cloud}, edge models face significant challenges in real incremental scenarios. Through the identification module, the system selectively uploads data to the cloud-based foundational model, powered by GPT-4, for queries, thereby enhancing motion planning performance. The inference process of the cloud model in real scenarios encompasses three critical stages: perception, prediction, and planning. These stages are essential for ensuring the model's efficient response.

\subsection{Edge-Cloud Collaborative Motion Planning}
This project utilizes data collected by autonomous vehicles at the Guangzhou International Campus of South China University of Technology as the testing benchmark. The dataset comprises images captured from the perspective of autonomous vehicles, with a lower camera angle that aligns closely with practical autonomous driving applications such as delivery and patrol.

\begin{figure}[!th]
    \centering
    \includegraphics[width=0.9\linewidth]{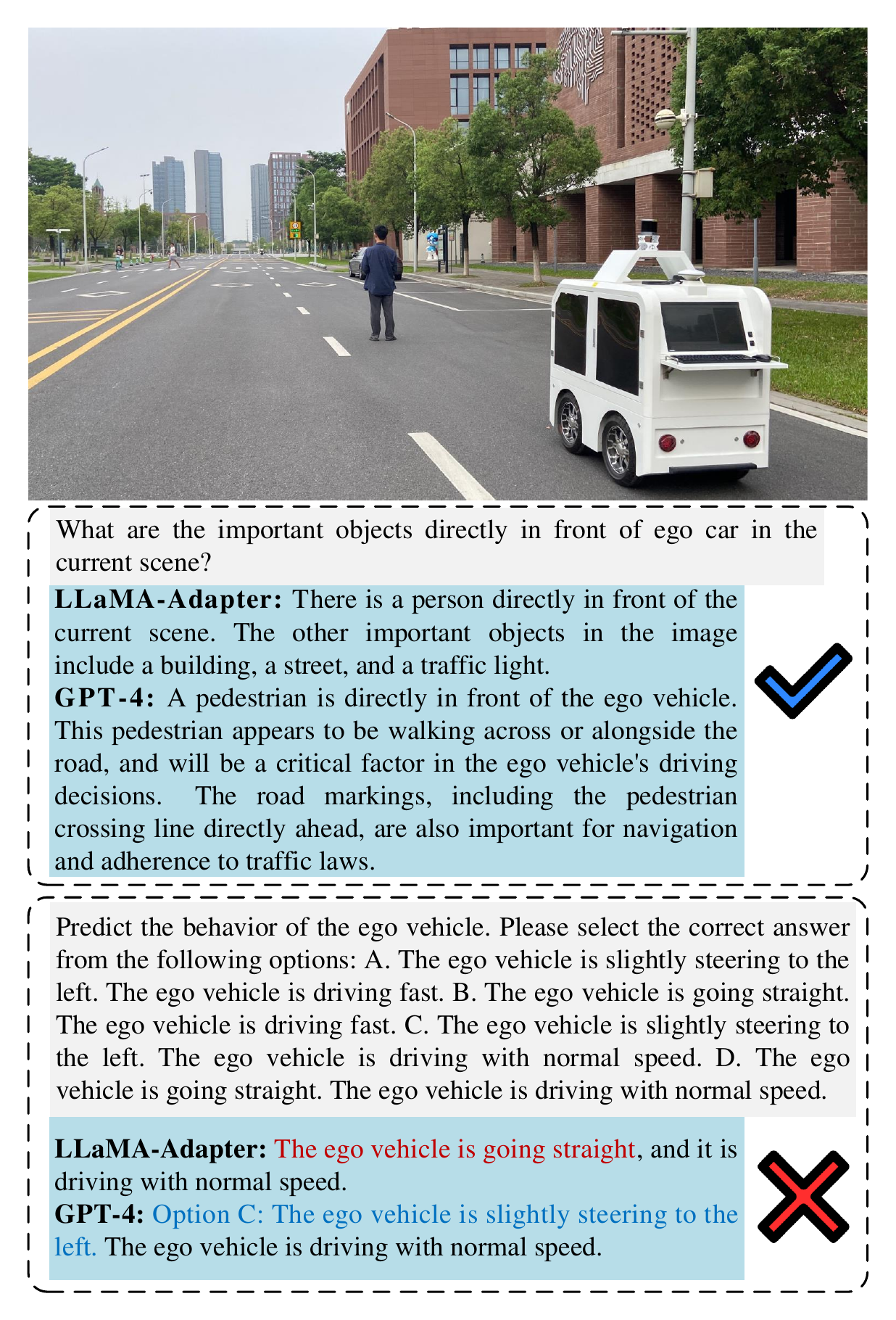}
    \caption{Comparison of inference results between edge and cloud models in the same scenario.}
    \label{fig:compare}
\vspace{-0.15in}
\end{figure}

Fig.~\ref{fig:compare} illustrates the inference outcomes of different models in the same scenario. In most cases, edge models (LLaMA-Adapter \cite{gao2023llama}) demonstrate performance comparable to cloud models (GPT-4 \cite{achiam2023gpt}), where invoking cloud models offers limited improvement to the driving task and may lead to resource wastage and unnecessary delays. 

Although the edge model is capable of making quick inferences in most cases, the cloud model demonstrates extremely high accuracy when dealing with complex scenarios. For instance, in pedestrian recognition and complex road planning (as shown in the second case of Fig.~\ref{fig:compare}), the cloud model can correct the inference errors made by the edge model, thereby enhancing the overall safety and reliability of the system.

\subsection{Further Analysis and Discussion}
Inspired by \cite{han2024dme}, we evaluate LLMs based on three metrics.
Gaze: Assessing the accuracy of LLMs in identifying areas of focus during the driving process. Scene Understanding: Evaluating the precision of LLMs in describing elements present in the current driving scene. Logic: Analyzing the correctness of the reasoning employed by LLMs in making driving decisions.

Tab.~\ref{tab:compare} presents the inference results of models of varying sizes within the dataset. The performance of cloud-based LLMs significantly surpasses that of edge-based small-scale models: As shown in the table, cloud-based LLMs (such as GPT-4 and GPT-4o) achieve higher scores across all three metrics (Gaze, Scene Understanding, and Logic) compared to edge-based small-scale models. Specifically, GPT-4 scores 87.1 in Gaze and 88.9 in Scene Understanding, significantly outperforming the highest scores of edge-based models, which are 66.8 and 59.4, respectively. \looseness=-1

Edge-based small-scale LLMs exhibit advantages in specific scenarios: Despite the superior overall performance of cloud-based LLMs, edge-based small-scale models demonstrate significant benefits in environments with limited computational resources or where low-latency responses are required. For instance, edge-based models such as Phi-2-2.7B and TinyLlama-1.1B provide relatively stable performance under constrained resources. \looseness=-1

\begin{table}
\caption{Performance comparison of edge and cloud models in autonomous driving, focusing on relevant driving metrics.}
\centering
\resizebox{1.0\linewidth}{!}{
\begin{tabular}{llccc}
\toprule
\multicolumn{1}{c}{Type} & \multicolumn{1}{c}{LLM} & Gaze ($\uparrow$) & \begin{tabular}[c]{@{}c@{}}Scene\\ Understanding ($\uparrow$)\end{tabular} & Logic ($\uparrow$) \\ \toprule
Edge                       & Moondream               & 54.7 & 52.6                                                          & 49.6  \\
Edge                   & OpenELM-450M \cite{mehta2024openelm}            & 59.5 & 52.1                                                          & 50.5  \\
Edge                   & TinyLlama-1.1B \cite{zhang2024tinyllama}          & 61.7 & 53.9                                                          & 54.1  \\
Edge                 & Gemma-2B                & 65.5 & 58.6                                                          & 59.9  \\
Edge                 & Phi-2-2.7B              & 66.8 & 59.4                                                          & 61.2  \\ 
Cloud                       & LLava-7B                & 72.3 & 74.2                                                          & 61.5  \\
Cloud                       & LLama-Adapter           & 75.1 & 79.4                                                          & 69.6  \\
Cloud                       & GPT-4o                  & 85.3 & 86.5                                                          & 80.3  \\
Cloud                       & GPT-4                  & 87.1 & 88.9                                                          & 81.6  \\\bottomrule
\end{tabular}
\label{tab:compare}
}
\vspace{-0.15in}
\end{table}

\section{Conclusion}
This study extensively investigates the application performance of LLMs in autonomous driving systems, leveraging edge computing, cloud computing, and edge-cloud collaborative processing. In the edge computing environment, the system swiftly processes real-time driving data, utilizing CoT for logical inference and decision-making. The cloud model exhibits exceptional perception, prediction, and planning capabilities when handling complex driving scenarios. Notably, the edge-cloud collaboration selectively uploads critical data to the cloud, not only enhancing inference speed and conserving communication resources but also significantly reducing system latency. This collaboration also markedly improves the edge model's understanding of incremental and complex scenarios, thereby enhancing overall system performance in motion planning. 
The experimental results validate the effectiveness and efficiency of the model in practical applications. 
These findings provide crucial theoretical and practical guidance for the future development of autonomous driving technologies.

\bibliographystyle{IEEEtran}  
\bibliography{IEEEabrv,references}
\end{document}